\documentclass{article}

\usepackage{arxiv}

\usepackage[utf8]{inputenc} 
\usepackage[T1]{fontenc}    
\usepackage{hyperref}       
\usepackage{url}            
\usepackage{booktabs}       
\usepackage{amsfonts}       
\usepackage{nicefrac}       
\usepackage{microtype}      
\usepackage{lipsum}		
\usepackage{graphicx}

\usepackage{natbib}
\usepackage{algorithm}
\usepackage{algpseudocode}
\usepackage[inline]{enumitem}
\usepackage{subcaption}
\usepackage{enumitem}
\usepackage{amsfonts}

\title{Competitiveness of MAP-Elites against Proximal Policy Optimization on locomotion tasks in deterministic simulations}

\date{} 					

\author{Szymon Brych\\
  Department of Computing\\
  Imperial College London\\
  London, SW7 2AZ, UK\\
  \texttt{szymon.brych@gmail.com} \\
  \And
  Antoine Cully\\
  Department of Computing\\
  Imperial College London\\
  London, SW7 2AZ, UK\\
  \texttt{a.cully@imperial.ac.uk} \\
}

\date{}

\renewcommand{\headeright}{}
\renewcommand{\undertitle}{}

\hypersetup{
pdftitle={Competitiveness of MAP-Elites against Proximal Policy Optimization on locomotion tasks in deterministic simulations},
pdfsubject={AI, ML, RL},
pdfauthor={Szymon Brych, Antoine Cully},
pdfkeywords={Quality-Diversity optimization, Reinforcement Learning, Proximal Policy Optimization, MAP-Elites},
    colorlinks=true,
    linkcolor=blue,
    filecolor=magenta,      
    urlcolor=blue,
    citecolor=black
}

\begin{document}
\maketitle

\begin{abstract}
The increasing importance of robots and automation creates a demand for learnable controllers which can be obtained through various approaches such as Evolutionary Algorithms (EAs) or Reinforcement Learning (RL). 
Unfortunately, these two families of algorithms have mainly developed independently and there are only few works comparing modern EAs with deep RL algorithms.
We show that Multidimensional Archive of Phenotypic Elites (MAP-Elites), which is a modern EA, can deliver better-performing solutions than one of the state-of-the-art RL methods, Proximal Policy Optimization (PPO) in the generation of locomotion controllers for a simulated hexapod robot.
Additionally, extensive hyper-parameter tuning shows that MAP-Elites displays greater robustness across seeds and hyper-parameter sets.
Generally, this paper demonstrates that EAs combined with modern computational resources display promising characteristics and have the potential to contribute to the state-of-the-art in controller learning.
\end{abstract}


\section{Introduction}


Increased demand for robots and various forms of automatic control creates a demand for automated ways to generate controllers. Such controllers were traditionally designed by experts, however, associated complexities make this process slow and therefore expensive. For this reason, it is particularly desirable to automatically learn controllers as opposed to manually designing them.

In this space, learning approaches seem particularly promising. Existing work has shown that Evolutionary Algorithms (EAs) can be successfully applied to control problems in common simulated environments \citep{petroski2017neuroevolution,salimans2017evolution}, as well as in real robots \citep{cully2015robots}. Particularly interesting among EAs is a family of Quality-Diversity algorithms (QDs; \cite{cully2017quality}). These take inspiration from the diversity of species produced by natural evolution. Instead of generating a single solution, like most learning algorithms, QD algorithms produce a collection of diverse and high-performing solutions. This diversity allows robustness to changing conditions as well as specialization to various use cases resulting in performance gains~\cite{kaushik2020adaptive}.

At the same time, Reinforcement Learning (RL) has gained huge popularity due to breakthroughs \citep{mnih2013dqn,silver2017mastering} fueled with the advent of wide-spread applications of deep neural networks \citep{DBLP:journals/corr/Schmidhuber14}. Over a relatively short course of time Deep RL has also seen numerous very impressive contributions to the problems of continuous control in simulation \citep{ddpg_algorithm,schulman2017proximal, heess2017emergence}.


These examples show that both EAs and RL methods can be applied in controller learning and despite slightly different terminologies they share the same high-level learning principles. Reinforcement Learning focuses on learning from experience through a \emph{reward signal} that is feedback on the desirability of actions taken with respect to the environment. In the context of EAs, the reward signal is often referred to as the \emph{fitness}, which in evolutionary terms quantifies how well an individual is adapted to the environment in which it acts. 

We attempt to make further contributions to the knowledge base of Evolutionary Algorithms and their competitiveness against Deep Reinforcement Learning approaches. To this end, we discuss the most relevant Policy Gradient methods \citep{williams1992simple, sutton2000policy} and choose Proximal Policy Optimization (PPO; \cite{schulman2017proximal}) for deeper investigation. Analogously, we further proceed with a discussion of some of recent EAs and select a QD algorithm MAP-Elites  \citep{mouret2015illuminating} for the comparison. We then specify a continuous gait control problem for an 18-degrees-of-freedom \emph{hexapod} robot in a deterministic simulation. Evaluations on a physical robot is left for future works. Subsequently, throughout series of experiments for two different setups of an open and closed-loop controller, we perform a systematic comparison of selected algorithms to eventually discuss  characteristics of both approaches.


The main contributions of this work are: 
\begin{enumerate*}
    \item {showing that MAP-Elites is a competitive alternative to PPO, achieving better results in terms of episode reward in all the evaluated scenarios.}
    \item {demonstrating greater robustness of MAP-Elites against both seeds and hyper-parameters, which is crucial when considering applying algorithms on physical robots as this results in better reproducibility and shorter tuning procedure.}
\end{enumerate*}

\section{Related work}
Learning locomotion controllers is a problem that has attracted the attention of multiple research domains. This is illustrated by the diversity of approaches that have been explored in the literature, for instance: Policy Gradient Methods \citep{twin_delayed_DDPG,mnih2016asynchronous,schulman2015trust,schulman2017proximal,ddpg_theory}, Evolutionary Algorithms \citep{cully2015robots,petroski2017neuroevolution,mouret2015illuminating}, Particle Swarm Optimization \citep{zambrano2013pso} or Bayesian Optimization \citep{calandra2016bayesian}. Additionally,  \cite{deisenroth2013survey} and \cite{calandra2016bayesian} give a great high-level comparison of aforementioned categories augmented with Grid Search and Random Search, which are often used as baseline benchmarks for algorithms \citep{calandra2016bayesian,cully2015robots,petroski2017neuroevolution}. In this work, we focus exclusively on the first 2 categories: Policy Gradient Methods and Evolutionary Algorithms.




\subsection{Policy Gradient Methods}
All Reinforcement Learning methods seek to optimize the objective $J(\pi)$ which is defined as the expected sum of future rewards $r_t$ discounted with a factor $\gamma$ under a trajectory distribution determined by following policy $\pi$ over a certain horizon $T$:
\begin{equation}
J(\pi)=\mathbb{E}_{\pi}\left[\sum_{t=0}^{T}\gamma^{t}r_{t}\right]
\end{equation}
Policy Gradient methods~\citep{williams1992simple,sutton2000policy} leverage a parametrized policy function $\pi_{\theta}$, such as a neural network, that is trained by optimizing an objective function with respect to policy parameters $\theta$. Additionally, it is quite common to approximate a state-value function $V(s)$:
\begin{equation}
    V(s)=\mathbb{E}_{\pi}\left[\sum_{k=t}^{T}\gamma^{k-t}r_{t}\middle|s_{t}=s\right]
\end{equation}
State-value function serves as a baseline when judging the utility of taking certain action in a given state and helps in stabilizing learning \cite{sutton2000policy}. Using this baseline, objective $J(\pi)$ can be optimized by a direct gradient estimation:
\begin{equation}
    \nabla_{\theta}J(\pi_{\theta})=\mathbb{E}_{\pi_{\theta}}\left[\sum_{t=0}^{T}\gamma^{t}\nabla_{\theta}log\pi_\theta(a_t|s_t)\left(\sum_{k=t}^{T}\gamma^{k-t}r_{t}-V(s_{t})\right)\right]
\end{equation}
One of the more recent improvements to PG methods is a Trust Region Policy Optimization (TRPO) \citep{schulman2015trust}, which introduced a new surrogate objective function used to limit the size of the policy's update (therefore "trust region"). This is designed to ensure that the policy does not diverge to unknow regions of the search space.
The introduction of TRPO has been followed by multiple influential works, such as in the Asynchronous Advantage Actor-Critic (A3C) algorithm \citep{mnih2016asynchronous} and Proximal Policy Optimization (PPO). A3C  combines the Actor-Critic PG approach \citep{sutton2018reinforcement} with parallel and asynchronous execution which added the power of modern computing to variants of existing algorithms.

On the other hand, PPO~\citep{schulman2017proximal} avoids the challenge of using a constrained optimisation by replacing TRPO's core objective component with a \emph{clipping objective}:
\begin{equation}
    L^{clip}(\theta)={\hat{\mathbb{E}}}_{t}[min(r_{t}(\theta)\hat{A}_{t},clip(r_{t}(\theta),1-\epsilon,1+\epsilon)\hat{A}_{t})]
\end{equation}
where $\hat{A}$ is an \emph{advantage estimator} at timestep $t$ (more details in \cite{schulman2015gae}), $r_{t}$ is a policy ratio $\pi_{\theta}(a_{t}|s_{t})/\pi_{\theta_{old}}(a_{t}|s_{t})$ and, the expectation stands for "the empirical average over a finite batch of samples" \cite{schulman2017proximal}. This component caps the incentives for altering parameters of the current policy if taking an  action according to a new policy $\pi_{\theta}$ differs too much in probability from the $\pi_{\theta_{old}}$ used for experience collection (for more details please refer to Alg.~\ref{algo:ppo}). This greatly simplifies the implementation, which combined with PPO's capabilities for parallelization, makes it a particularly attractive tool. Additionally, its applicability to learning locomotion has already been reported in several publications \citep{schulman2017proximal,heess2017emergence}. 

\subsection{Evolutionary Algorithms}
Another family of algorithms suitable for use in learning locomotion controllers is the family of Evolutionary Algorithms (EAs). Since we are interested in comparisons to Deep Reinforcement Learning, we focus on evolving controllers represented as neural networks. The approach of evolving neural networks is sometimes termed as a \emph{Neuroevolution} (e.g., in \cite{petroski2017neuroevolution}). The idea of using evolutionary approaches to learn controllers is not new and has been discussed in numerous publications across the years  \citep{reeve2005analysis,lee2013hyperneat,cully2015robots,salimans2017evolution,petroski2017neuroevolution,huan2018adaptive}. 

Some of the more recent discussions of EAs include work presented in \cite{salimans2017evolution}, which uses a gradient-based algorithm called Evolutionary Strategy. This algorithm defines a population of parameters as a Gaussian distribution, which is sampled to approximate the gradient of the fitness function. Then, stochastic gradient descent is used to update the parameters of the Gaussian distribution, which moves the population to regions of the search space with high-fitness. Authors highlight strong parallelization capabilities and successfully train their controllers in both continuous and discrete domains.

Contrary to the aforementioned, authors of \cite{petroski2017neuroevolution} offer an implementation of an EA algorithm simply referred to as the Genetic Algorithm, which doesn't rely on gradients. It maintains a population of individuals that are subject to the typical evolutionary operations of fitness evaluation, selection (with elitism), and mutation. No crossover is applied. One of the most distinguishing aspects of this work is the scale, as individuals take the form of networks having up to 4M parameters.

\href{quality-diversity.github.io}{Quality-Diversity optimisation} (QD) algorithms is a recently introduced subgroup of EAs \citep{cully2017quality} that aim at producing large collections of solutions which are both diverse and high-performing. One of the most well-known QD algorithms is the Multidimensional Archive of Phenotypic Elites (MAP-Elites; \cite{mouret2015illuminating, cully2015robots}). It defines an archive with a fixed number of cells that is used to store the collection of solutions. Each cell corresponds to a different type of solution that is characterised by a different \emph{behavioral descriptor}. The behavioural descriptor is a numerical vector describing some task-dependent behavioral properties of an individual (e.g., the average body oscillations, the end location of the robot or the leg-ground contact patterns). The objective of this paper is to evaluate how MAP-Elites performs compared to PPO on the generation of locomotion controller for a hexapod robot. 



\section{Methodology}
\subsection{Common setup}
In this work, we compare PPO and MAP-Elites on locomotion tasks. In order to ensure a fair comparison, a common simulation environment setup is used across experiments. It consists of a 3D model of a 6-legged, hexapod robot with 18 actuated degrees-of-freedom (3 per leg). This model is designed to resemble a physical robot used in \citep{cully2015robots}.

The locomotion task is the same throughout this paper - it is to walk as far as possible along an X-axis within an episode lasting 5 seconds (corresponding to 333 time steps). This walked distance is referred to as the \emph{fitness} or \emph{episode reward}.
The environment is deterministic (i.e., the same action always leads to the same result) and simulated with the open-source physics simulator DART \citep{lee2018dart}. 

\subsection{Algorithms and implementation}

\paragraph{PPO} The pseudo-code of PPO~\cite{schulman2017proximal} is presented in Alg.~\ref{algo:ppo}. During each iteration, several actors generate batches of experience by following the policy $\pi_{\theta_{old}}$. This is then followed by the computation of advantage estimators $\hat{A}$. Subsequently, the objective $L$, which includes terms for: clipping $L^{c}$, squared loss $L^{v}$ and entropy $S$ (with hyper-parameters: $\epsilon$, $c_1$, $c_2$), is optimized with respect to the current policy's parameters $\theta$ over K epochs. Each epoch utilizes experience split into several mini-batches. After this is done, the old policy's parameters get overwritten with current ones and the procedure repeats or learned policy is returned.



\paragraph{MAP-Elites} Alg.~\ref{algo:mapelites} shows the pseudo-code of MAP-Elites~\citep{mouret2015illuminating,cully2015robots}, where $(\mathcal{P}, \mathcal{C})$ stands for a behavior-performance archive being a result of execution. During each generation, we sample and mutate several solutions from the behavior-performance archive. The algorithm then evaluates each of these mutated solutions and records their behavioral descriptors and performance based on results from a single episode (rollout) per individual. These new solutions are then potentially added in the archive if a cell indicated by the behavioral descriptor is empty or occupied by a lower-performing solution.

\begin{minipage}{0.53\textwidth}
\begin{algorithm}[H]
\small
\caption{PPO Algorithm \citep{schulman2017proximal}}
\label{algo:ppo}
\begin{algorithmic}
\For{iteration $  = 1,2,...$} 
\For{actor $  = 1\to N$} 
\State $ $Run policy $\pi_{\theta_{old}}$ in env for T timesteps
\State $ $Compute advantage estimates $\hat{A}_{1}$,...,$\hat{A}_{T}$%
\EndFor
\State $ $Optimize $L$ with respect to $\theta$, 
\State $ $   with K epochs and mini-batch size M $\leq$ NT where:
\State $ $    $L_{t}(\theta)={\hat{\mathbb{E}}}_{t}[L_{t}^{c}(\theta)+c_{1}L_{t}^{v}(\theta)+c_{2}S[\pi_{\theta}](s_{t})]$
\State $ $   $L_{t}^{c}(\theta)=min(r_{t}(\theta)\hat{A}_{t},clip(r_{t}(\theta),1-\epsilon,1+\epsilon)\hat{A}_{t})$
\State $ $   $L_{t}^{v}(\theta)=(V_{\theta}(s_{t})-V^{targ})^{2}$
\State $ $   $r_{t}(\theta)=\pi_{\theta}(a_{t}|s_{t})/\pi_{\theta_{old}}(a_{t}|s_{t})$
\State $ \theta_{old} \leftarrow
\theta $
\EndFor
\State \Return policy $\pi_{\theta_{old}}$
\end{algorithmic}
\end{algorithm}
\end{minipage}
\hfill
\begin{minipage}{0.46\textwidth}
\begin{algorithm}[H]
\small
\caption{MAP-Elites Algorithm \citep{mouret2015illuminating}}
\label{algo:mapelites}
\begin{algorithmic}
\State $(\mathcal{P} \leftarrow \emptyset, \mathcal{C} \leftarrow \emptyset)$
\For{iter $  = 1\to I$} 
\If{iter $< N$} 
  \State $\mathbf{c'}\leftarrow $ random\_controller()   
\Else 
  \State $\mathbf{c}\leftarrow $ random\_selection($\mathcal{C}$) 
  \State $\mathbf{c'}\leftarrow $ random\_variation($\mathbf{c}$) 
\EndIf
\State $\mathbf{x'}\leftarrow $behavioral\_descriptor(simu($\mathbf{c'}$)) 
\State $p'\leftarrow $performance(simu($\mathbf{c'}$)) 
\If{$\mathcal{P}(\mathbf{x'})= \emptyset$ or $\mathcal{P}(\mathbf{x'})<p'$}
\State $\mathcal{P}(\mathbf{x'})\leftarrow p'$ 
\State $\mathcal{C}(\mathbf{x'})\leftarrow \mathbf{c'}$ 
\EndIf
\EndFor
\State \Return behavior-performance map ($\mathcal{P}$ and $\mathcal{C}$)
\end{algorithmic}
\end{algorithm}
\end{minipage}

All of the MAP-Elites experiments in this work use a discrete, 6-dimensional behavior descriptor, recording fraction of the total time that a given leg of the hexapod was in the contact with the ground \citep{cully2015robots}. Each dimension of a descriptor has four or five buckets (\emph{descriptor base}).
The source code of both \href{https://github.com/Antymon/ppo_cpp}{PPO} and \href{https://gitlab.doc.ic.ac.uk/sb5817/neural-mape}{MAP-Elites} is available under respective repositories.

\subsection{Comparisons}

Due to the different nature of PPO and MAP-Elites, there are some difficulties associated with their comparison. To name some of the high-level differences:
\begin{itemize*}
    \item PPO, like other Policy Gradient methods, optimizes for the expected cumulative reward from an episode for a single solution, whereas MAP-Elites has an additional goal of producing a diverse set of solutions according to a behavioral descriptor.
    \item PPO outputs stochastic policies that drive its exploration. On the other hand, MAP-Elites, as used in this work, produces deterministic policies that are decoupled from exploration mechanics, as the exploration is driven by mutation.
    \item PPO gets transition and reward information at the frame-level granularity, whereas MAP-Elites only receives cumulative rewards (i.e., fitness) from episodes. However, MAP-Elites additionally uses behavioral descriptors, which quantify the high-level behavioral features and PPO has no access to such information.
\end{itemize*}

Nevertheless, there are still some aspects of both algorithms that make quantitative comparisons possible:
\begin{itemize*}[leftmargin=*]
    \item Their total computational effort can be measured in the number of simulation frames.
    \item Solutions yielded by PPO can be compared with the best-fit individual of MAP-Elites in terms of cumulative episodic rewards.
\end{itemize*}

To make comparisons between MAP-Elites and PPO as fair as possible the following measures are taken:
\begin{itemize*}[leftmargin=*]
    \item MAP-Elites is set up to evolve the weights of neural policies of the same size as the network architectures used by PPO (value network excluded as this was treated as PPO's implementation detail irrelevant to MAP-Elites).
    \item The main policy network architecture is selected to be a fully-connected Multi-Layer Perceptron, although this is a limiting choice for MAP-Elites which is capable of simultaneously evolving the weights and topologies of neural networks.
    \item To minimize differences in the setup, the original Python implementation of PPO is ported to C++ where it can directly use the same instance of the simulation environment and robot model as MAP-Elites does.
\end{itemize*}

\section{Experimental evaluation}

Two sets of experiments are presented: 
\begin{enumerate*}[leftmargin=*]
    \item \emph{Open-loop gait training} - a scenario where the policy takes only a function of time as an input.
    \item \emph{Closed-loop gait training} - a more complex scenario where the state of the robot's body is the policy's input. 
\end{enumerate*}

The comparisons of the different evaluations are based on statistics over the episode reward (i.e., cumulative episode reward for a fixed amount of simulation frames).
We perform multiple replication runs to examine seed sensitivity due to stochastic components in the tested algorithms. The statistical significance of performance difference is verified through \emph{Wilcoxon Signed-Rank Test} and \emph{p-values} are reported. 
\begin{figure}
\centering
    \includegraphics{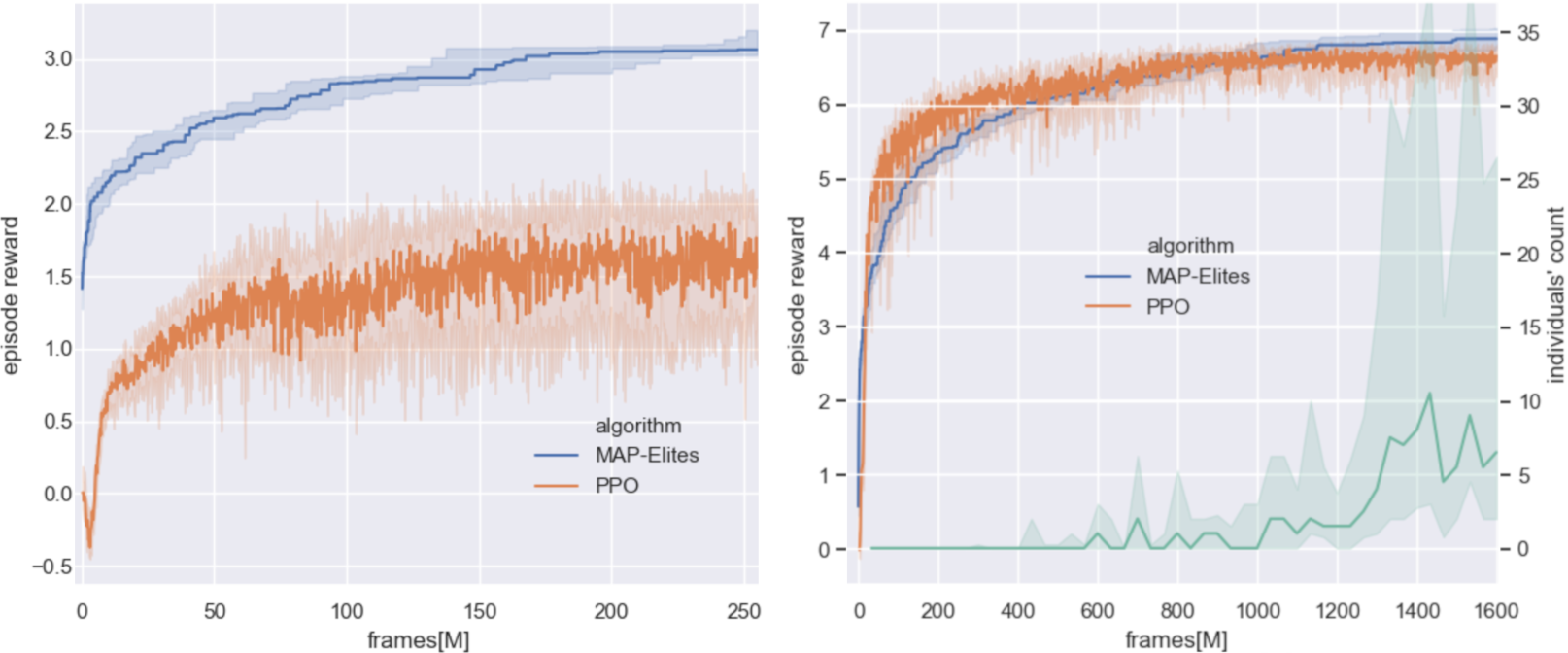} 
\caption{Comparison of the quartile analysis for the open-loop (left) and closed-loop (right) experiments featuring PPO and MAP-Elites. A colorful shading around a line represents the variability between the first and the third quartile, while the thicker line is the median. For the closed-loop, apart from episode rewards, a distribution over the number of individuals in MAP-Elites archives which over-perform the median of PPO is showed in green.
}
\label{fig:quartiles}
\end{figure}

\subsection{Open-loop Controller}

The open-loop controllers studied in this work take a scalar \emph{time-modulo-period} as an input and output 18 target angular positions of joints. This is quite popular due to its high utility and simplicity \citep{cully2015robots}.
The period is always set arbitrarily to one second. 

All of the used open-loop architectures have two layers of hidden neurons with two to six neurons in each layer. These counts are treated as a hyper-parameter. Biases were used, which results in total policy parameter count ranging from 64 to 130.

\subsubsection{PPO hyper-parameter tuning}

The hyper-parameter selection is done in two phases. First, 370 unique PPO hyper-parameter configurations are tested for a short horizon (75M frames) with four replications each. Then, the four best configurations (according to the median episode reward\footnote{We report medians as it seems to be a better-suited statistic than for instance average, due to higher robustness against various distributions. An analogous argument applies to the preference of the inter-quartile range over standard deviation.}) are executed for a longer horizon (255M frames) with 20 replications each. We report the performance of the single best performing configuration on the long horizon in Fig.~\ref{fig:quartiles} (left). The details of the hyper-parameter values are given in Table \ref{hp_ppo}.
The reasoning behind this is to initially evaluate hyper-parameter configurations until the first horizon, and continue only with the most promising ones. This approach allows to explore a larger number of hyper-parameter configuration at a lower computational cost. 

The hyper-parameter configurations sampled for the initial assessment consist of learning rate: [5e-5,1e-2] and clipping range: [5e-2,4e-1], both sampled log-uniformly. An entropy term is sampled log-uniformly from the range [1e-4,1e-2] in 25\% cases or set to zero otherwise in the remaining 75\%. Additionally, among uniformly sampled hyper-parameters are mini-batch size, selected from the range of [2,32] Ki, and policy architecture sampled from 8 predefined options of [64,130] parameters in each. Both policy and value networks have the same architecture, but weights are not shared.

\subsubsection{MAP-Elites hyper-parameter tuning}

The hyper-parameter tuning procedure for MAP-Elites is the same as the one used with PPO. However, due to MAP-Elites being far more stable across parameters and substantially dominating in performance (Fig.~\ref{fig:quartiles}) we test 220 hyper-parameter configurations with three replications for initial horizon and, as before, 20 replications for the best ones. The performance of a MAP-Elites algorithm is defined as the episode reward of the best solution contained in the Behavioural-Performance archive.
The hyper-parameters range for MAP-Elites include: mutation rate sampled uniformly from range [0,0.5], the same set of neural architectures as for PPO, and selection of either base-4 or base-5 behavioral descriptor. The details of the hyper-parameter values for the selected configuration are given in Table \ref{hp_mape}.



\subsubsection{Open-loop results}

Fig.~\ref{fig:quartiles} (left) shows the experimental results with the median, first and third quartile displayed. We can see that MAP-Elites with MLP individuals significantly overperforms PPO (medians of 3 vs 1.5 meters traveled by the hexapod). This difference is statistically significant with a p-value of 0.00014.
Additionally, we can also see that MAP-Elites tends to increase the performance over longer horizons than PPO.
The best-fit individual after the first generation performs above 1.5 meters. That is much better than many parametrizations of PPO were able to reach throughout the whole of their run showing the importance of a relatively simple random search, which was also appreciated in \citep{petroski2017neuroevolution}.

Qualitatively, gaits learned through the open-loop PPO training quite often take a form of a hexapod flipping over to the back and then walking upside-down (e.g., \href{https://youtu.be/c6ddhHfHFsI?t=7}{"tip-over" gait}), which can be thought of as a local optimum. 
The viability of this gait relates to specifics of the hexapod model, which has lower leg extents above knee joints.
On the other hand, gait referred to as \href{https://youtu.be/c6ddhHfHFsI?t=13}{"aspiring bipedal"} is an example of a gait produced by MAP-Elites.
\subsection{Closed-loop Controller}


In the closed-loop setting, both the input and output consist of 18 angular positions of the joints. The input is the current state of the robot, while the output refers to the desired state for the next time step. Similar setup can be found in \citep{heess2017emergence}. An increased number of inputs results in an increased number of parameters overall. The predefined set of architectures ranges from 98 to 282 in the number of weights (including biases).

\subsubsection{Hyper-parameter tuning}
In the closed-loop case, we follow the same evaluation scheme as before, except that for both MAP-Elites and PPO, we sample only 50 hyper-parameter configurations with  three replications each over the short horizon. Then, we select the best-performing configuration and perform 20 replications over the long horizon. In this experience, the short horizon is 533M frames and the long one 1.6B frames.

The sampling distributions of PPO hyper-parameters are just as before with exception of batch sizes selected from just 3 options of 16, 32, or 64Ki as inspired by similar settings of \citep{heess2017emergence}. In the case of MAP-Elites, the hyper-parameters were sampled just as in the open-loop controller setting.  The details of the final hyper-parameter values are given in Table \ref{hp_ppo}.

\begin{figure}
\centering
\begin{subfigure}{.5\textwidth}
    \includegraphics{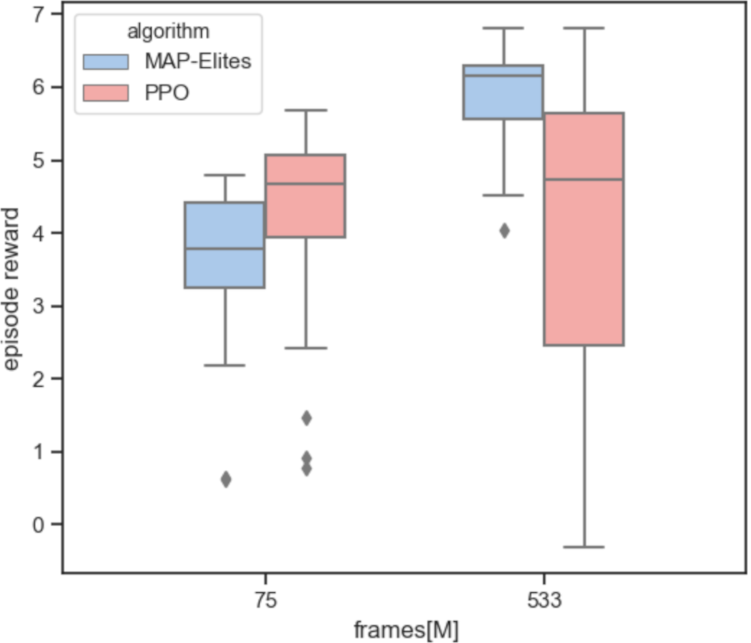} 

\end{subfigure}
    \begin{subfigure}{.42\textwidth}
    \vspace*{-0.6cm} 
    \includegraphics{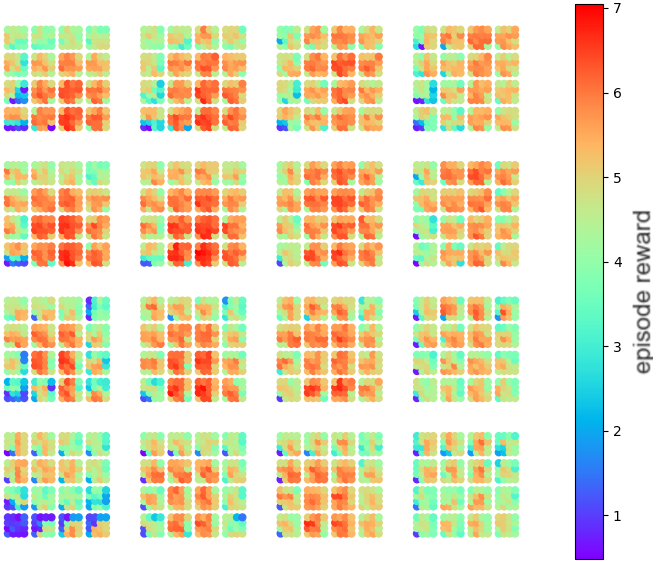} 
    \end{subfigure}
        \caption{Left: Aggregations of medians of performance across batches of configurations. Each box plot consists of closed-loop configurations and aggregates medians of performance per configuration from three replications after the selected horizon. Right: An example of the last archive for the closed-loop MAP-Elites configuration from Table~\ref{hp_mape}. It features 4096 individuals where color indicates a return from an episode (the performance). Three levels of squares correspond to six dimensions of behavioral descriptor: one per each leg of the hexapod. Each dimension has four buckets capturing a fraction of the episode's total time steps that a given leg was in contact with the ground.}
\label{fig:closed_loop_all}
\end{figure}


\subsubsection{Closed-loop results}
The statistics over the single best configurations' runs of each algorithm are presented in Fig.~\ref{fig:quartiles} (right). In the closed-loop setting, PPO is a much more competitive than in the open-loop setting. As Fig.~\ref{fig:quartiles} (right) shows, PPO quickly increases performance in terms of simulation frames, however, it eventually flattens out at a median performance of around 6.6 meters.
The characteristics of the gaits produced by PPO changed dramatically in this scenario, delivering very fast gaits (e.g., \href{https://youtu.be/c6ddhHfHFsI?t=24}{"gallop"} gait).

In contrast, MAP-Elites' performance ascends over a much longer horizon and converges at a higher performance median value of around 6.9 meters without the risk of divergence. The p-value yielded by Wilcoxon Signed-Rank Test is 6e-4, therefore confirming the statistical difference between algorithms' results at the end of the horizon.
It is important to recall that MAP-Elites does not provide just one policy, but instead a large collection of diverse and high-performing solutions. Fig.~\ref{fig:quartiles} (right) also shows the number of controllers in the MAP-Elites archives that outperform the median performance of PPO at each frame. We can see that MAP-Elites produces between 3 to 25 outperforming controllers, for the same computational budget as PPO. Gait named \href{https://youtu.be/c6ddhHfHFsI?t=29}{"tiptoe"} is an example of a closed-loop controller evolved with MAP-Elites.

Additionally, in Fig.~\ref{fig:closed_loop_all} (left) we analyze the sensitivity of both algorithms to hyper-parameter selection. We can see that for the relatively short horizon of 75M frames, median of median performances across configurations is higher for PPO than MAP-Elites and the interquartile range is similar. After 533M frames, PPO results across configurations display increased inter-quartile range and similar median as in shorter runs. This is not the case for the MAP-Elites configurations which reduce their variability and substantially increase the median to the extent which makes it dominating over corresponding PPO configurations. Fig.~\ref{fig:closed_loop_all} (right) shows an example archive for the best-performing MAP-Elites configuration and randomly selected run after total of 1.6B time steps with the best individuals in red.

\begin{table}[ht]
\begin{tabular}{  r | l  l  } 
 &  Open-loop & Closed-loop  \\ 
 \hline
clip\_range ($\epsilon$) & 0.08458 & 0.27059 \\ 
learning\_rate & 2.7175e-3 & 1.52e-5 \\ 
layer\_0\_size & 3 & 5\\
layer\_1\_size & 4 & 5\\
entropy ($c_{2}$) & 1.4938e-4 & 0 \\ 
\end{tabular}
\quad
\begin{tabular}{  r | l  l  } 
 &  Open-loop & Closed-loop  \\ 
 \hline
nb\_gen & 3825 & 24001 \\ 
mutation\_rate & 0.0891977 & 0.188637 \\ 
layer\_0\_size & 4 & 4 \\ 
layer\_1\_size & 4 & 5 \\ 
descriptor\_base & 4 & 4 \\ 
\end{tabular}
\small
\center{
\caption{PPO (left) and MAP-Elites (right) final hyper-parameter configurations selected after tuning. Additionally, both PPO experiment setups used 10 epochs, 32 mini-batches, batch size of 32Ki, discount factor $\gamma=0.99$ and General Advantage Estimator \cite{schulman2015gae} factor $\lambda=0.95$.}
\label{hp_ppo}
\label{hp_mape}
}
\end{table}

\section{Discussion}

The behavior of PPO in terms of the open-loop gait controller learning seems quite contrary to MAP-Elites. PPO was subject to oscillations and drops during the training, showed less robustness to hyper-parameters and seeds (especially for small batch size).
Aforementioned \href{https://youtu.be/c6ddhHfHFsI?t=7}{"tip-over" gait}, found in the course of the open-loop PPO experimentation, can be thought of as easy-to-find local optimum.
On the other hand, MAP-Elites demonstrated great exploration capabilities, without a need for additional stimulation like the entropy term of PPO~\cite{heess2017emergence}. MAP-Elites was able not only to double the median episode return value after a certain computational horizon but also found such an original gait as \href{https://youtu.be/c6ddhHfHFsI?t=13}{"aspiring bipedal"}.
These results are rather surprising as MAP-Elites, optimizing for two goals: diversity and performance, seems disadvantaged by design when compared with a method that solely focuses on performance. 

Despite nearly doubled search effort in configuration count to find a well-performing open-loop configuration of PPO, none of them were statistically competitive against MAP-Elites. However, despite more than 1000 runs, PPO's hyperparameter space is of very high dimensionality, which makes it still very possible that more impressive configurations do exist. 
One must also remember that the discussed open-loop setting is 
deterministic (stochastic injection into MAP-Elites proves to be troublesome \citep{stochastic_mape,flageat2020fast}).


In the case of the closed-loop setting, the performance gap between algorithms proved to be less obvious. 
PPO seems to prioritize exploitation which allows for quicker ascend, wheres MAP-Elites needs a longer horizon to dominate over a contender. This behavior is a well-known property of MAP-Elites as its exploration capability may result in a long time to find a high-performing solution, which in our case is also the better-performing solution.

Generally, due to PPO's greater number of hyperparameters and the sensitivity to their actual values (exemplified in Fig.~\ref{fig:closed_loop_all}), it tends to be more difficult to find stable configurations, even despite the insights of similar applications in the literature \citep{heess2017emergence} and available software \citep{MuJoCo}.  For this reason, throughout whole the experimentation more effort was spent on the examination of PPO (PPO:620 vs MAP-Elites:500 billion frames total). This seemed a reasonable course of action given our objective of providing a fair comparison and that PPO tended to systematically show lower performance than MAP-Elites.

MAP-Elites by design has the characteristics of non-decreasing performance for the best-fit individual. Thanks to this feature, learning is stable without oscillations or drops, which are typical to the RL methods like PPO. Since PPO maintains only one solution that is constantly altered, it may also have a problem with reverting undesirable actions, whereas MAP-Elites would simply discard an inefficient individual within the context of the certain behavioral bin.




\section{Conclusions and future work}
This work presents a thorough comparison of an Evolutionary Algorithm MAP-Elites and Policy Gradient method PPO in the context of robotic gait learning in deterministic simulation. To achieve statistically-significant results we perform numerous replications which combined with hyper-parameter search require a computational effort of around one trillion simulation frames.

Our work tends to suggest that MAP-Elites, if applicable, might be easier to implement in a parallel setting and tends to deliver slightly better results over the long horizon even when compared with top Policy Gradient approach like PPO. At the same time, robustness to seed and hyper-parameter selection proves very convenient when working with MAP-Elites, saving compute that would be otherwise invested in these selection activities, and encouraging reproducibility.

Although due to the fairness of comparison, we constrain MAP-Elites to evolve neural networks with a fixed-topology, this is unnecessary in the general case. Throughout the experimentation, we have also obtained some preliminary results on evolving neural network topology altogether with weights. These results hinted at further improvements of performance, producing another original gait such as \href{https://youtu.be/c6ddhHfHFsI?t=20}{"bunny hop" gait}. This idea could be further investigated as well as considering hybrid approaches or addressing the problem of MAP-Elites with noisy observations.




\section{Acknowledgements}

We would like to thank Marek Barwiński, Ben Cataldo, Luca Grillotti, and Joe Phillips for their efforts in reading the paper draft and their valuable comments.

\bibliographystyle{unsrt}


\end{document}